\documentclass[runningheads]{llncs}

\usepackage{graphicx}

\usepackage{algorithm}
\usepackage[noend]{algpseudocode}
\usepackage{url}
\usepackage{listings}
\usepackage{wrapfig}
\usepackage{caption}
\usepackage{subfig}

\DeclareRobustCommand{\hllime}[1]{{\sethlcolor{lime}\hl{#1}}}

\DeclareUnicodeCharacter{0301}{\'{e}}

\usepackage[T1]{fontenc}

\usepackage{soulutf8}
\newcommand{\sameThanks}{\textsuperscript{\thefootnote}}
\usepackage{microtype}

\usepackage{todonotes}

\newcommand{\titleacronym}{DAHRS}

\begin{document}

\title{DAHRS: Divergence-Aware Hallucination-Remediated SRL Projection}

\titlerunning{Divergence-Aware Hallucination-Remediated SRL Projection}
%
\author{Sangpil Youm\inst{1}\thanks{Corresponding Author.}
\and Brodie Mather\inst{2}
\and Chathuri Jayaweera\inst{1},\newline
Juliana Prada\inst{1} \and Bonnie Dorr\inst{1}
}

%
\authorrunning{Youm et al.}
%
\institute{University of Florida, Gainesville, FL, USA \email{\{youms\sameThanks,chathuri.jayawee,bonniejdorr,juliana.prada\}@ufl.edu}\and IHMC, Pensacola, FL, USA \email{bmather@ihmc.org} 
}

%
\maketitle              
%
\begin{abstract}
Semantic role labeling (SRL) enriches
many downstream applications, e.g., machine translation, question answering, summarization, and stance/belief detection. However, building multilingual SRL models is challenging due to the scarcity of semantically annotated corpora for multiple languages. Moreover, state-of-the-art SRL projection (XSRL) based on large language models (LLMs) yields output that is riddled with spurious role labels. Remediation of such hallucinations is not straightforward due to the lack of explainability of LLMs. We show that hallucinated role labels are related to naturally occurring divergence types that interfere with initial alignments. We implement \textit{Divergence-Aware Hallucination-Remediated SRL projection} (\titleacronym{}), leveraging linguistically-informed alignment remediation followed by greedy \textit{First-Come First-Assign} (FCFA) SRL projection. \titleacronym{} improves the accuracy of SRL projection without additional transformer-based machinery, beating XSRL in both human and automatic comparisons, and advancing beyond headwords to accommodate phrase-level SRL projection (e.g., EN-FR, EN-ES). Using CoNLL-2009 as our ground truth, we achieve a higher word-level F1 over XSRL: 87.6\% vs. 77.3\% (EN-FR) and 89.0\% vs. 82.7\% (EN-ES). Human phrase-level assessments yield 89.1\% (EN-FR) and 91.0\% (EN-ES). We also define a divergence metric to adapt our approach to other language pairs (e.g., English-Tagalog). 


\end{abstract}





\vspace*{-.2in}
\section{Introduction}
\label{sec:intro}
\vspace*{-.1in}

The natural language processing (NLP) task of semantic role labeling (SRL) captures ``\textit{who did what to whom}'' for many downstream applications, e.g.,
machine translation, question answering, and summarization \cite{Liu2010SemanticTranslation,Genest2011FrameworkGeneration}. Semantic roles are 
central to 
inferring unstated information (e.g., stances \cite{Mather2021ADetection,Mather2022FromDomains} and emotional cues \cite{Campagnano2022SRL4EFramework})
that are absent from the output of
NLP tools such as dependency parsing.

Disappointingly, SRL has been studied primarily in English due to highly available English-specific SRL annotated datasets \cite{Fei2020Cross-LingualCorpus}. The scarcity of multilingual SRL-annotated corpora motivates the need for cross-language approaches that project semantic roles from English to other languages.

Many studies have explored pre-trained SRL models \cite{Mehta2018TowardsLabeling,Shi2019SimpleLabeling} and generative AI approaches for semantic tasks that include SRL \cite{tsai2023llamaloop}. These LLM-centric studies tend to focus exclusively on English.
The associated LLMs thus introduce hallucinations without obvious recourse due to an inherent lack of explainability.

Our approach, ``Divergence-Aware Hallucination-Remediated SRL Projection'' (\titleacronym{}) adopts a generalized characterization of divergence types 
\cite{Dorr1994MachineSolution,Maniyar2021LinguisticPerceptive} and corrects alignnments, remediating hallucinated semantic-role transfer from source to target languages (e.g., English-French and English-Spanish). We introduce a greedy ``First-Come First-Assign'' (FCFA) algorithm within \titleacronym{} that projects roles from corrected initial alignments. FCFA also remediates the hallucinated lack of semantic role projections emerging from corrected initial alignments.

The key insight here is that leveraging linguistic knowledge overcomes deficiencies in current transformer-based alignment-projection approaches. Transformer-based alignment treats target words as a bag-of-words, frequently aligning source-language terms to hallucinated target-language terms. By contrast, \titleacronym{} injects an awareness of naturally occurring language \textit{divergences}, e.g., one-to-many/many-to-one translations or word/phrase order distinctions, into alignment. Straightforward correction of alignments that would otherwise lead to hallucinated \textit{incorrect} roles supports effective and explainable transfer of semantic roles from the source language to the target language. 

State-of-the-art XSRL \cite{Daza2020X-SRL:Dataset} addresses a subset of language divergences explored in this paper: nominalizations and separable verb prefixes. In cases where the initial alignment is correct, XSRL fails to project valid roles in the context of other types of divergences, often hallucinating a \textit{lack} of semantic role projections on the right-hand side. 
\titleacronym{} is designed to address two types of hallucinations simultaneously: alignment and projection. The performance of \titleacronym{} is compared to that of XSRL using data processed by both methods (see section~\ref{sec:results}). 

Hallucination remediation in \titleacronym{} starts with token-level and phrase-level corrections to an initial transformer-based mBERT \cite{Dou2021WordCorpora} alignment. Following this, additional hallucination remediation takes place during projection. Fig.~\ref{fig:divergence} illustrates two representative cases of \textit{divergences} that have triggered hallucinations in prior work: \textit{Light Verb} and \textit{Structural}.\footnote{Fig.~\ref{fig:divergence} inputs: (a) EN:  The dow 's dive was the 12th - worst ever and the sharpest since the market fell 156.83 FR: La chute du dow jones a été la 12e - la pire et la plus forte depuis que le marché a chuté de 156.83. (b) EN: Some ``circuit breakers'' installed after the october 1987 crash failed their first test. FR: Certains ``disjoncteurs'' installés après l'écrasement d'octobre 1987 ont échoué leur premier test.} Square brackets `[]' indicate SRL projections, with unaligned words indicated by $\epsilon$. The output shown at each stage explainably pin-points which sub-components fail or succeed (alignment or projection, or both). 



\textbf{(a) Light Verb Divergence.} The single verb \textit{fell} maps to a combination of a ``light'' verb (\textit{a}) and content word ``fallen'' (\textit{chuté}).  Despite the correct initial mBERT alignment, XSRL is unable to ``see past'' this divergence to project semantic roles to the target-language side. The inherent uninterpretability of the underlying models impedes the ability to determine what has gone awry, but we observe that this divergence type almost leads to a hallucinated \textit{lack} of SRL assignments. By contrast, \titleacronym{} correctly transfers labels V, ARG1 (EN \textit{market} to FR \textit{marché}), and ARG2 (EN \textit{156.83} to FR \textit{156.83}), leaving \textit{a, de} appropriately unassigned. Also, \textit{chuté} is an adjectival participle in French, but its verbal nature supports ARG1 assignments, so the V label is retained by design. 

\begin{wrapfigure}{r}{0.55\textwidth}
\vspace*{-.25in}
\scriptsize

\textbf{(a) Light Verb (Hallucinated \textit{lack} of roles):}\\ \textcolor{purple}{\textit{market fell 156.83}} -
\textcolor{blue}{\textit{marché \textbf{a chuté} de 156.83}}

mBERT-based Alignment:
\vspace*{-.1in}
\begin{flushleft}
\begin{tabular}{ l c l }
\mbox{~~}\textcolor{purple}{market} & --- & \textcolor{blue}{marché} \\ 
\mbox{~~}{$\epsilon$} & --- & \textcolor{blue}{a} \\ 
\mbox{~~}\textcolor{purple}{fell} & --- & \textcolor{blue}{chuté} \\ 
\mbox{~~}{$\epsilon$} & --- & \textcolor{blue}{de} \\ 
\mbox{~~}\textcolor{purple}{156.83} & --- & \textcolor{blue}{156.83} \\
\end{tabular}
\end{flushleft}
\vspace*{-.1in}

XSRL:
\vspace*{-.1in}
\begin{flushleft}
\begin{tabular}{ l c l }
\mbox{~~}\textcolor{purple}{[ARG1] market} & --- & \textcolor{blue}{marché} \\ 
\mbox{~~}{$\epsilon$} & --- & \textcolor{blue}{a} \\ 
\mbox{~~}\textcolor{purple}{[V] fell} & --- & \textcolor{blue}{chuté} \\ 
\mbox{~~}{$\epsilon$} & --- & \textcolor{blue}{de} \\ 
\mbox{~~}\textcolor{purple}{[ARG2] 156.83} & --- & \textcolor{blue}{156.83} \\
\end{tabular}
\end{flushleft}
\vspace*{-.1in}

\titleacronym{}:
\vspace*{-.1in}
\begin{flushleft}
\begin{tabular}{ l c l }
\mbox{~~}\textcolor{purple}{[ARG1] market} & --- & \textcolor{blue}{[ARG1] marché} \\ 
\mbox{~~}{$\epsilon$} & --- & \textcolor{blue}{a} \\ 
\mbox{~~}\textcolor{purple}{[V] fell} & --- & \textcolor{blue}{[V] chuté} \\ 
\mbox{~~}{$\epsilon$} & --- & \textcolor{blue}{de} \\ 
\mbox{~~}\textcolor{purple}{[ARG2] 156.83} & --- & \textcolor{blue}{[ARG2] 156.83} \\
\end{tabular}
\end{flushleft}
\vspace*{-.1in}

\vspace*{0.05in}
\textbf{(b) Structural (Hallucinated \textit{incorrect} roles):}\\ \textcolor{purple}{\textit{october 1987 crash}} -\textcolor{blue}{\textit{écrasement d' octobre 1987}}

mBERT-based Alignment:
\vspace*{-.1in}
\begin{flushleft}
\begin{tabular}{ l c l }

\mbox{~~}\textcolor{purple}{october} & --- & \textcolor{blue}{\'ecrasement} \\ 
\mbox{~~}\textcolor{purple}{october} & --- & \textcolor{blue}{octobre} \\ 
\mbox{~~}{$\epsilon$} & --- & \textcolor{blue}{d'} \\ 
\mbox{~~}\textcolor{purple}{1987} & --- & \textcolor{blue}{1987} \\ 
\mbox{~~}\textcolor{purple}{crash} & --- & \textcolor{blue}{\`ecrasement}
\end{tabular}
\end{flushleft}
\vspace*{-.1in}
XSRL:
\vspace*{-.1in}
\begin{flushleft}
\begin{tabular}{ l c l }
\mbox{~~}\textcolor{purple}{[ARGM-TMP] october} & --- & \textcolor{blue}{[ARGM-TMP] octobre} \\ 
\mbox{~~}{$\epsilon$} & --- & \textcolor{blue}{d'} \\ 
\mbox{~~}\textcolor{purple}{[ARGM-TMP] 1987} & --- & \textcolor{blue}{[ARGM-TMP] 1987} \\ 
\mbox{~~}\textcolor{purple}{[ARGM-TMP] crash} & --- & \textcolor{blue}{$\epsilon$} \\ 
\end{tabular}
\end{flushleft}
\vspace*{-.1in}

\titleacronym{}:
\vspace*{-.2in}
\begin{flushleft}
\begin{tabular}{ lcl }
\mbox{~~}\textcolor{purple}{[ARGM-TMP] october} & --- & \textcolor{blue}{[ARGM-TMP] octobre} \\ 
\mbox{~~}{$\epsilon$} & --- & \textcolor{blue}{d'} \\ 
\mbox{~~}\textcolor{purple}{[ARGM-TMP] 1987} & --- & \textcolor{blue}{[ARGM-TMP] 1987} \\ 
\mbox{~~}\textcolor{purple}{[ARGM-TMP] crash} & --- & \textcolor{blue}{[ARGM-TMP] \'ecrasement}
\end{tabular}
\end{flushleft}

\vspace*{-.2in}
\caption{Divergence cases corresponding to two hallucination types: (a) Light Verbs introduce one-to-many/many-to-one divergences that impede XSRL transfer of semantic roles even when the initial alignment is correct, thus hallucinating a \textit{lack} of roles on the target-language side; (b) Structural divergences introduce word/phrase order distinctions that result in extra, spuriously aligned terms, thus hallucinating \textit{incorrect} roles.}
\label{fig:divergence}
\vspace*{-.3in}
\end{wrapfigure}

\textbf{(b) Structural Divergence.} 
A difference in source/target word order 
(\textit{October 1987 crash} vs. \textit{crash of October 1987})
combined with a bag-of-words design leads to an incorrect mBERT alignment. Here, \textit{October} aligns to \textit{Octobre} (October) and a (hallucinated) occurrence of \textit{\'ecrasement}, while \textit{crash} aligns to a second occurrence of \textit{\'ecrasement}. The resulting XSRL projection includes \textit{incorrect} role transfers, leaving \textit{crash} unaligned and thus without a role. By contrast,
\titleacronym{} applies alignment remediation, mapping \textit{crash} to \textit{écrasement}, and \textit{October} to \textit{Octobre}, and correctly transferring ARGM-TMP to French.

\titleacronym{} identifies divergence types, remediates hallucinations at both the token/phrase level, and applies greedy FCFA SRL projection. Divergence handling couples alignment remediation with FCFA, which is parameterized to include syntactic properties of the source language (e.g., English is head-initial) to accommodate proper SRL projection. This simple, efficient design transcends ``yet another transformer'' in both accuracy and explainability.

While numerous studies have focused on improving explainability in diverse NLP tasks and applications such as classification~\cite{Liu2020TowardsClassification} 
or medical NLP~\cite{Danilevsky2020AProcessing}
, to our knowledge, ours is the first to address explainability for SRL in NLP. Our visualization of alignment and projection decisions
(see Fig.~\ref{fig:divergence}) displays accessible, linguistically relevant representations associated with SRL transfers (and predicates, indicated as ``V'').
These visualized linguistic representations display how and why each SRL projection is made, highlighting the handling of translation divergences throughout the entire 
process.


Below we present related work, followed by a description of \titleacronym{}. We then present automated and human-validated evaluations. We demonstrate that \titleacronym{} outperforms XSRL in 
accuracy (87.6\% vs. 77.3\% F1 (EN-FR), 89.0\% vs. 82.7\% F1 (EN-ES)).
We discuss the potential for generalization to low-resource languages. We then conclude and explore future work.

\vspace*{-.2in}
\section{Related Work}
\label{sec:relatedwork}
\vspace*{-.14in}

Early applications for annotation-projection include: 
dependency parsing \cite{Hwa2005BootstrappingTexts}; part-of-speech taggers \cite{Yarowsky2001InducingCorpora}; machine translation \cite{Zhang2008AModel,Shen2016Cross-languageTranslation}; divergence-inspired alignment \cite{Dorr2002DUSTer:Alignment}; and creation of syntactic-dependency datasets for multiple languages \cite{McDonald2013UniversalParsing}. We borrow the notion of annotation projection to produce explainable, cross-language SRL that advances the state of the art. 

A contrasting SRL annotation projection 
approach 
is one where a source-language model is modified 
for direct applicability
to a new language, using cross-lingually shared representations \cite{Kozhevnikov2013Cross-lingualModels}. Such ``model transferring'' approaches do not 
align datasets across languages, but instead induce a separate dataset. By contrast, annotation projection approaches (including our own) propagate available information from one language to another via alignment.

Translation-based models provide an alternative approach for transferring SRL annotations.
These
have demonstrated promising performance due to recent improvements in neural machine translation (NMT)
\cite{Fei2020Cross-LingualCorpus,Gehring2017ConvolutionalLearning,Hassan2018AchievingTranslation}. Translation-based projection involves tree-to-tree mappings to build cross-lingual SRL-annotated corpora \cite{Prazak2017Cross-LingualDependencies}, based on tree/graph-based representations \cite{Shen2016Cross-languageTranslation}. By contrast, our approach aims to accommodate divergences for SRL projection via word-to-word mapping without relying on additional structure (e.g., trees or graphs).

Prior studies have demonstrated the benefits of embedding models in cross-language SRL projection. 
For example, Polyglot SRL \cite{Mulcaire2018PolyglotLabeling} employs word vectors and is trained on the union of annotations between two languages. A cross-lingual encoder-decoder model is applied to simultaneously translate and apply SRL for resource-poor languages \cite{Daza2019TranslateLabeling}. Adding a syntactic information layer to the embedding models demonstrates plausibility of transferring semantic roles \cite{Guarasci2022}. By contrast, our approach enables improved SRL projection without additional vector-based machinery. 
Instead, we factor out syntactic variations, as these are not central to the transfer of semantic roles, and introduce a greedy SRL projection algorithm that is both accurate and efficient.

Translation divergences and associated alignment errors
lead to considerable noise, 
often resulting in the implementation of intricate techniques.
For example, projection probability distributions and gold-standard annotated data have been employed to improve alignment performance \cite{Akbik2015GeneratingLabeling}. 
XSRL
uses 
translations produced by DeepL \cite{DeepL}, more than 
10\% of 
which
are human-judged as improperly translated and 
removed. 
An mBERT \cite{Devlin2018BERT:Understanding} aligner is applied,
followed by an additional transformer-based mechanism (BERT Score) \cite{Zhang2019BERTScore:BERT}, to project semantic roles to the target sentence. 
Although these approaches offer valuable SRL projection
strategies,
two major concerns are 
the added complexity (e.g., BERT-based scoring)
and, in the case of XSRL, 
human filtering to remove noisy translations. The latter
negatively impacts the resulting training data coverage.

While our approach involves projection, it differs from those above in that it operates on all translated sentence pairs (no human filtering) and produces a greedily induced
SRL projection.
The resulting annotations
are consistent with translation divergence studies.
Decisions on projected labels are made readily accessible and easily visualized, rather than hidden behind \textit{black box} algorithms. 

\vspace*{-.15in}
\section{Divergence-Aware Hallucination-Remediated SRL Projection (\titleacronym{})}
\label{sec:methodology}
\vspace*{-.1in}

\begin{figure}
\vspace*{-.3in}
  \includegraphics[width=0.99\textwidth,
                    height=2.5cm
                    ]
  {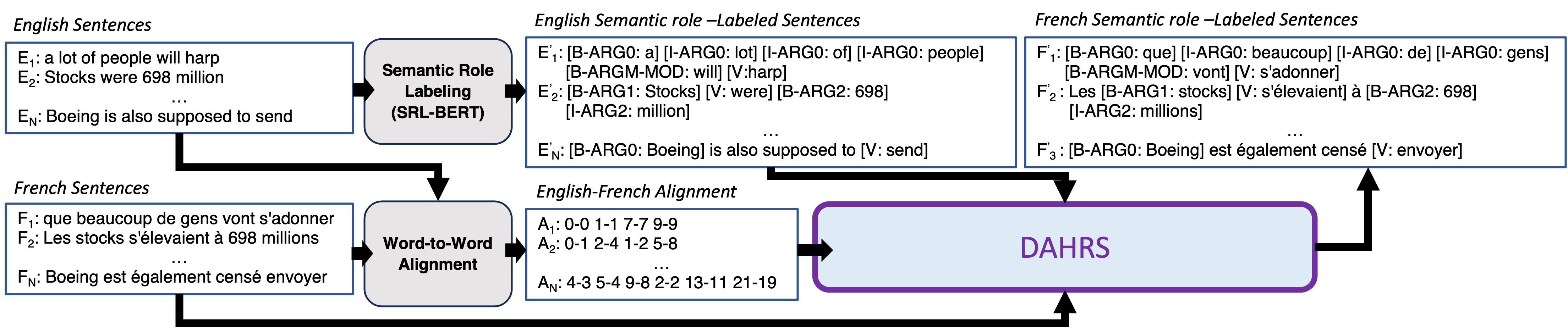}
  \vspace*{-.1in}
  \caption{
  \textit{Divergence-Aware Hallucination-Remediated SRL Projection} (\titleacronym{}) pipeline from English to French}
\label{fig:flowchart}
\vspace*{-.25in}
\end{figure}

\titleacronym{}'s key contribution is its ability to compensate for potential semantic role errors emerging from hallucinated alignments that coincide with naturally occurring cross-language divergences. Leveraging source-language knowledge (e.g., English is head initial) coupled with a greedy FCFA algorithm, \titleacronym{} transfers semantic roles to the target language. 


\begin{wrapfigure}{r}{0.5\textwidth}

\centering
\vspace*{-.4in}
  \includegraphics[width=6cm,
                    height=1.7cm
                    ]
  {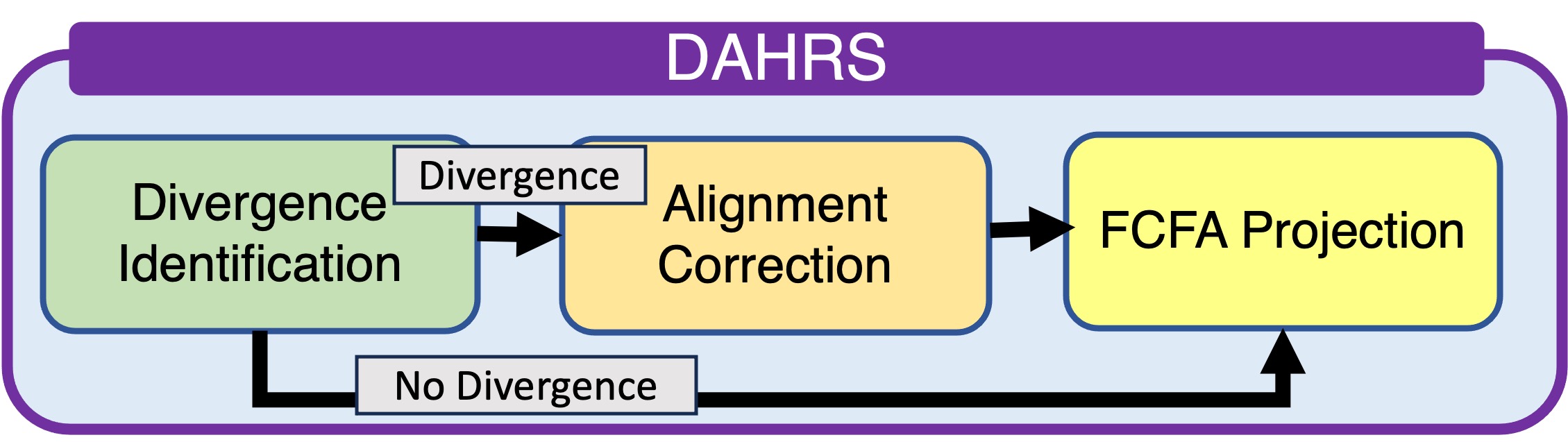}
  \caption{Divergence-Aware Hallucination Remediated SRL Projection (DAHRS)}
\label{fig:DAHRS_essential}
\vspace*{-.32in}
\end{wrapfigure}

Fig.~\ref{fig:flowchart} illustrates the \titleacronym{} step-wise pipeline with an English-to-French example. \titleacronym{}'s input is an initial mBERT-style alignment, as in XSRL, but prior to SRL projection it corrects hallucinated alignments and transfers semantic roles without additional transformer-based processing.

Fig.~\ref{fig:DAHRS_essential}
shows three key steps in \titleacronym{}:
divergence identification (see Section~\ref{sec:div-id}), alignment correction (\titleacronym{}$_1$ and \titleacronym{}$_2$, see Section~\ref{sec:DAHRS_Algorithm}), and FCFA projection (\titleacronym{}$_3$, also in Section~\ref{sec:DAHRS_Algorithm}). 
When divergence identification uncovers a divergence, \titleacronym{} modifies the alignment prior to SRL projection. Otherwise it directly projects semantic roles through FCFA projection.


\vspace*{-.1in}
\subsection{Divergence Identification}
\label{sec:div-id}
\vspace*{-.05in}
For divergence identification, \titleacronym{} relies on a sub-categorization of divergences into three types, as shown in Fig.~\ref{fig:divergence_e}. For example, with regard to 
the divergences illustrated in Section~\ref{sec:intro}, \textit{Light Verb} divergences are associated with (a) one-to-many and (b) many-to-one sub-categories, and \textit{Structural} divergences are associated with (c) the ordering sub-category. 

\begin{wrapfigure}{r}{0.35\textwidth}
\vspace*{-.03in}

\scriptsize{
\begin{flushleft}
\begin{tabular}{ l c l }
\multicolumn{3}{l}{\textbf{(a) One-to-many}}\\
\textcolor{purple}{laptops} & ------ & \textcolor{blue}{ordinateurs} \textcolor{black}{; 17-23} \\ 
\textcolor{purple}{laptops} & ------ & \textcolor{blue}{portables} \textcolor{black}{; 17-24}\\ 
\end{tabular}
\end{flushleft}
}
\vspace*{-.1in}
\scriptsize{
\vspace*{-.15in}
\begin{flushleft}
\begin{tabular}{ l c l }
\multicolumn{3}{l}{\textbf{(b) Many-to-one}}\\
\textcolor{purple}{fell} & ------ & \textcolor{blue}{effondrée} \textcolor{black}{; 4-6}\\ 
\textcolor{purple}{apart} & ------ & \textcolor{blue}{effondrée} \textcolor{black}{; 5-6} 
\end{tabular}
\end{flushleft}
}

\vspace*{-.3in}
\scriptsize{
\begin{flushleft}
\begin{tabular}{ l c l }
\multicolumn{3}{l}{\textbf{(c) Ordering}}\\
\textcolor{purple}{october} & ------ & \textcolor{blue}{\'ecrasement} \textcolor{black}{; 9-7}  \\ 
\textcolor{purple}{october} & ------ & \textcolor{blue}{octobre} \textcolor{black}{; 9-9}  \\ 
{$\epsilon$} & ------ & \textcolor{blue}{d'} \textcolor{black}{; $\epsilon$-8}  \\ 
\textcolor{purple}{1987} & ------ & \textcolor{blue}{1987} \textcolor{black}{; 10-10}  \\ 
\textcolor{purple}{crash} & ------ & \textcolor{blue}{\`ecrasement} \textcolor{black}{; 11-7}  \\ 
\end{tabular}
\end{flushleft}
}
\vspace*{-.25in}
\caption{Three subcategories of divergences (token level): One-to-many, Many-to-one, and Ordering}
\label{fig:divergence_e}
\vspace*{-.3in}
\end{wrapfigure}
The identification of these divergence sub-categories for a given source-target input pair relies on position-value pairs. These pairs indicate the tokens and phrases that are mapped singularly or repeatedly across the source and target inputs.
Divergence 
types are identified across tokenized
source and target sentences, where
each 
token 
is assigned a position value starting from 0. 

Consider the French sentence fragment
\textit{ordinateurs portable} (\textit{laptops})  in Fig.~\ref{fig:divergence_e}(a). This string is associated with position values of 17 in English and 23,24 in French. 
Source and target word mappings are denoted by a hyphenated position-value pair. For example, 17-23 and 17-24 indicate the 17th English word (\textit{laptops}) aligns with the 23rd and 24th word French words (\textit{ordinateurs portable}).
This case is identified as a one-to-many divergence, i.e., a single source token aligns with multiple target tokens. 
Analogously, a many-to-one divergence is identified when multiple source tokens align with a single target token, as in Fig.~\ref{fig:divergence_e}(b), where \textit{fell}(4) and \textit{apart}(5) align with \textit{effondr\'ee}(6).

An ordering divergence is detected when a single source token is mBERT-aligned with multiple target tokens (one-to-many) while one of those same target tokens aligns with a different source token (many-to-one). Returning to our earlier example, \textit{October 1987 crash} (translated in French as \textit{crash of October 1987}), as shown in Fig.~\ref{fig:divergence_e}(c): \textit{october}(9) aligns with \textit{\'ecrasement}(7) and \textit{octobre}(9), while one of target tokens, \textit{\'ecrasement}(7) also aligns with \textit{crash}(11).

Although state-of-the-art (mBERT-based) word-to-word alignment establishes a reasonable source-to-target baseline, ordering divergences are not adequately handled, due to mBERT's bag-of-words design. These lead to incorrect alignments that must be remediated in order to avoid hallucinated SRL projections. We note that ordering distinctions have been a focus in statistical machine translation (SMT) for quite some time \cite{Rottmann2007WordModel}, but these have heretofore not been remediated for projection. 

Subsequent to identifying divergence types, as described below, our approach remediates hallucinations due to divergences and projects semantic roles through FCFA SRL projection.


\vspace*{-.15in}
\subsection{\titleacronym{} Algorithms}
\label{sec:DAHRS_Algorithm}
\vspace*{-.048in}

\titleacronym{}'s three key steps each correspond to a component-level algorithm: alignment correction at the token level (Algorithm~\ref{alg:fcfaalgo_token}) and phrase level (Algorithm~\ref{alg:fcfaalgo_phrase}) to remediate hallucinated \textit{incorrect} role projections, followed by FCFA SRL projection (Algorithm~\ref{alg:fcfaalgo_projection}) which remediates hallucinated \textit{lack} of role projections.

\vspace*{-.1in}
\subsubsection{$\titleacronym{}_1$: Token-Level Hallucination Remediation.}

We remediate alignment hallucinations at the token level, using $\titleacronym{}_1$ (see Algorithm~\ref{alg:fcfaalgo_token}). Such hallucinations are discerned from 
input 
pairs for one-to-one (\textit{tLevelOneToOne}), one-to-many (\textit{tLevelOneToMany}), many-to-one (\textit{tLevelManyToOne}) alignments. Additionally, 
a head-initial flag (\textit{headInitialFlag)} 
ensures proper SRL projection. This algorithm outputs a list of remediated alignments (\textit{remOneToOne}).

\begin{wrapfigure}{r}{0.59\textwidth}
\vspace*{-.5in}
    \begin{minipage}{0.59\textwidth}
        \begin{algorithm}[H]
        \caption{Token-level Hallucination Remediation ($\titleacronym{}_1$)}
        \label{alg:fcfaalgo_token}
        \scriptsize
        \vspace*{-.1in}
            \begin{flushleft}
            \textbf{Input} tLevelOneToOne, tLevelOneToMany, tLevelManyToOne, headInitialFlag\\
            \textbf{Output} remOneToOne\\
            \end{flushleft}
        \vspace*{-.15in}
        \begin{algorithmic}[1]
        \Function{$\titleacronym{}_1$}{tLevelOneToOne,tLevelOneToMany,
        tLevelManyToOne,headInitialFlag}
            \State remOnetoOne $\gets [ ]$
            \State remTargetWords $\gets [ ]$ 
            
            \For{$(src, tgt) \in \textit{tLevelOneToOne}$}
                \State\textit{remOneToOne}.insert((\textit{src}, \textit{tgt}))

            \EndFor
            
            \State tgtOneOne $\gets$ targets of tLevelOneToOne
            \For{$(src, tgtList) \in tLevelOneToMany$}
                \For{$tgt \in tgtList$}
                    \If {$tgt \in tgtOneOne$}
                        \State \textit{tgtList}.delete(\textit{tgt})
                    \Else
                    \State\textit{remOneToOne}.insert((\textit{src},\textit{tgt}))
                    \EndIf
                \EndFor
            \EndFor

            \State srcOneOne $\gets$ sources of tLevelOneToOne
            \For {\textit{(srcList,tgt)} $\in$ \textit{tLevelManyToOne}}
                \If {$src \in srcOneOne$}
                        \State \textit{srcList}.delete(\textit{src})
                \Else
                    \If {$\textit{headInitialFlag} \equiv\textit{True}$}
                        \State\textit{remOneToOne}.insert((\textit{srcList[0]},\textit{tgt}))
                     \Else
                        \State\textit{remOneToOne}.insert((\textit{srcList[1]},\textit{tgt}))
                    \EndIf
                \EndIf
            \EndFor
            \State\Return remOneToOne
        \EndFunction
        \end{algorithmic}
        \end{algorithm}
    \end{minipage}
\vspace*{-.25in}
\end{wrapfigure}

$\titleacronym{}_1$ initializes remediated alignments (\textit{remOneToOne}), inserting mBERT-aligned source-target token pairs specified in the \textit{tLevelOneToOne} list (lines 4-5). Next the target tokens in the one-to-many pair list (\textit{tLevelOneToMany}) are examined for alignment with other source tokens, preparing for hallucination remediation (line 6). If a target token is found to be aligned with an alternate source token, the hallucinated alignment is removed from the target token list (\textit{tgtList}) (lines 7-10). This action remediates alignment hallucinations that emerge in the context of ordering divergences. For example, in the earlier baseline alignment in Fig.~\ref{fig:divergence_e}(c), the word \textit{october} is incorrectly aligned with \textit{\'ecrasement}. This is detected due to the simultaneous \textit{october}-\textit{octobre} alignment (where no other source word aligns with \textit{octobre}). The spurious \textit{october}-\textit{\'ecrasement} alignment is 
hypothesized to be a hallucination and is removed.

After remedying spurious alignments in the one-to-many pairs, $\titleacronym{}_1$ proceeds to store the corrected 
source and target pairs in the output (\textit{remOneToOne}) (lines 11-12). In the earlier baseline alignment in Fig.~\ref{fig:divergence_e}(a), $\titleacronym{}_1$ correctly maps \textit{laptops} to both \textit{ordinateurs} and \textit{portables}. 

In the case of many-to-one alignment, $\titleacronym{}_1$ examines the source tokens in the one-to-one pair list (\textit{tLevelOneToOne}) for alignment with other target tokens, preparing for additional hallucination remediation (line 13). In this case, the algorithm addresses the potential for hallucinated (downstream) SRL projections due to the presence of particles or modifiers (e.g., \textit{apart} in \textit{fell apart}) that are aligned with the main verb. 

Remediation removes such tokens from the source token list (\textit{srcList}) (lines 14-16). For eaxample, in the earlier baseline alignment in Fig.~\ref{fig:divergence_e}(b), the \textit{apart}-\textit{effondrée} alignment is deleted. The remaining \textit{fell}-\textit{effondrée} alignment is retained and is positioned in the output (\textit{remOneToOne}) according to the \textit{headIntitalFlag}, where ``True'' indicates a head-initial language, selecting the first token and ``False'' indicates a head-final language (lines 17-21).

\vspace*{-.1in}
\subsubsection{$\titleacronym{}_2$: Phrase-Level Hallucination Remediation.}

$\titleacronym{}_2$ shown in Algorithm~\ref{alg:fcfaalgo_phrase} advances beyond the token-level processing of state-of-the-art (XSRL) in that it includes handling of phrases for SRL projection. 

\begin{wrapfigure}{r}{0.6\textwidth}
\vspace*{-.6in}
    \begin{minipage}{0.6\textwidth}
        \begin{algorithm}[H]
        \caption{Phrase-level Hallucination Remediation ($\titleacronym{}_2$)}
        \label{alg:fcfaalgo_phrase}
        \scriptsize
        \vspace*{-.14in}
            \begin{flushleft}
            \textbf{Input} pLevelOneToOne, pLevelOneToMany, pLevelManyToOne
            , srcPhRange, tgtPhRange, funcWordIdx, headInitialFlag \\
            \textbf{Output} remOneToOne\\
            \end{flushleft}
        \vspace*{-.15in}
        \begin{algorithmic}[1]
            \Function{$\titleacronym{}_2$}{pLevelOneToOne,pLevelOneToMany, pLevelManyToOne, srcPhRange, tgtPhRange, funcWordIdx, headInitialFlag}
        
                \If{{$ pLevelManyToOne = \emptyset$} and \newline
                \mbox{~~~~~~~~~~~~~}{$ pLevelOneToMany = \emptyset$}}
                    \State remOneToOne $\gets$ pLevelOneToOne
                    \State\Return remOneToOne
                \Else
                    \State tgtOneOne $\gets$ targets index of pLevelOneToOne
                    \For{$(src, tgtList) \in pLevelOneToMany$}
                        \For{$tgt \in tgtList$}
                            \If {\textit{tgt} $\notin$ \textit{tgtPhRange} $\mid$ \textit{tgt} $\in$ \textit{tgtPlevelOneOne}}
                                \State \textit{tgtList}.delete(\textit{tgt})
                            \EndIf
                        \EndFor
                        \For{$tgt \in tgtList$}             
                            \State\textit{remOneToOne}.insert((\textit{src},\textit{tgt}))
                        \EndFor
                    \EndFor
                    \State srcOneOne $\gets$ sources of pLevelOneToOne
                    \For{$(srcList, tgt) \in pLevelManyToOne$}
                        \For{$src \in srcList$}
                            \If {\textit{src} $\notin$ \textit{srcPhRange} $\mid$ \textit{src} $\in$ \textit{srcOneOne} $\mid$ \textit{src} $\in$ \textit{funcWordIdx}}
                                \State \textit{srcList}.delete(\textit{src})
                            \EndIf
                        \EndFor
                        
                        \If {size of \textit{srcList} $\equiv$ 1}
                            \State\textit{remOneToOne}.insert((\textit{srcList[0]},\textit{tgt}))
                        \Else
                            \If {$\textit{headInitialFlag} \equiv \textit{True}$}                          
                                \State\textit{remOneToOne}.insert((\textit{srcList[0]},\textit{tgt}))
                            \Else
                                \State\textit{remOneToOne}.insert((\textit{srcList[1]},\textit{tgt}))
                            \EndIf
                        \EndIf
                    \EndFor
                \EndIf
                \State\textit{remOneToOne} $\gets$\textit{remOneToOne} $+$ \textit{pLevelOneToOne}
                \State\Return remOneToOne
            \EndFunction
        \end{algorithmic}
        \end{algorithm}
    \end{minipage}
\vspace*{-.2in}
\end{wrapfigure}
Phrase-level processing is similar to what is described above, but phrase identification is employed: 
BIO (Begin-Inside-Outside) tags are assigned to the source-language side via SRL-BERT \cite{Shi2019SimpleLabeling}.\footnote{SRL-BERT achieves an F1 Score of 86.49 on the English Ontonotes dataset \cite{Weischedel2013OntoNotesLDC2013T19}, and it can be used non-exclusively. \url{https://allenai.org/terms}.} These BIO-delineated phrasal units are brought together with alignment corrections for more robust alignment hallucination remediation.
A phrase range is determined by arranging the source words in the order they appear within the sentence and employing BIO tags to identify phrases on the English side.\footnote{A phrase consists of a token that begins with a ``B'' tag and continues with tokens that have an ``I'' tag. The following token will have a new ``B'', an ``O'', or end of the sentence, indicating the end of the phrase.}

Phrase information (start to end indices), encoded as a
source phrase range (\textit{srcPhRange}) and target phrase range (\textit{tgtPhRange}), 
acts as phrase-level hallucination remediation input.
Other inputs are lists of phrase-level alignment pairs: one-to-one (\textit{pLevelOneToOne}), one-to-many (\textit{pLevelOneToMany}), many-to-one (\textit{pLevelManyToOne}). To support remediation, a list of function words (\textit{funcWordIdx}) and a head-initial flag (\textit{headInitialFlag}) are also introduced. This algorithm returns lists of remediated mappings (\textit{remOneToOne}). 

First, $\titleacronym{}_2$ examines whether the mBERT-aligned input is indicative of a one-to-many or many-to-one divergence within a given phrase (a BIO-tagged pair). If no such divergence is present, all the tokens in the phrase are returned as output (\textit{remOneToOne}) without correction (lines 2--4).

The next step remediates a detected hallucinated alignment resulting from an ordering divergence (lines 7--10). For each target list of phrasal one-to-many alignments, two aspects are examined: whether any target tokens are outside the corresponding phrase range, and whether any target tokens are simultaneously aligned with other source tokens. Tokens meeting one of these conditions are removed from the target list (\textit{tgtList}).
After $\titleacronym{}_2$ remediates spurious alignments in the one-to-many pair list, non-hallucinated source and target token pairs are stored in the output (\textit{remOneToOne}) (lines 11--12).

\begin{wrapfigure}{r}{0.53\textwidth}
\scriptsize
\vspace*{-.35in}
\begin{flushleft}
\begin{tabular}{ l c l }
\textcolor{purple}{[I-ARG1] circuit} & --- & \textcolor{blue}{$\epsilon$} \\ 
\textcolor{purple}{[I-ARG1] breakers} & --- & \textcolor{blue}{disjoncteurs}\textcolor{black}{; 4-2}\\
\textcolor{purple}{[B-V] installed} & --- & \textcolor{blue}{install\'es}\textcolor{black}{; 6-4}\\
\textcolor{purple}{[I-ARGM-TMP] after} & --- & \textcolor{blue}{apr\`es}\textcolor{black}{; 7-5}\\
\textcolor{purple}{[I-ARGM-TMP] the} & --- & \textcolor{blue}{l'} \textcolor{black}{; 8-6}\\ 
\textcolor{purple}{\hl{[I-ARGM-TMP] october}} & --- & \textcolor{blue}{\hl{écrasement}} \textcolor{black}{; 9-7}\\ 
\textcolor{purple}{\hl{[I-ARGM-TMP] october}} & --- & \textcolor{blue}{\hl{octobre}} \textcolor{black}{; 9-9}\\ 
\textcolor{purple}{[I-ARGM-TMP] 1987} & --- & \textcolor{blue}{1987} \textcolor{black}{; 10-10}\\ 
\textcolor{black}{\hllime{[I-ARGM-TMP] crash}} & --- & \textcolor{black}{\hllime{écrasement}} \textcolor{black}{; 11-7}\\ 
\end{tabular}
\end{flushleft}
\vspace*{-.2in}
\caption{One-to-many (\hl{yellow}) and Many-to-one (\textcolor{black}{\hllime{green}}) phrase-level alignments} 
\label{fig:manytomany-phrase}
\vspace*{-.25in}
\end{wrapfigure}

Lastly, $\titleacronym{}_2$ remediates hallucinated alignments arising from many-to-one divergences (lines 14-24). Three conditions are tested for each source token that aligns to a given target token: whether the source token is correctly located within corresponding range, whether it is aligned with another target token, and whether the source token is function word. Any source token matching one of those conditions is removed from the source list (\textit{srcList}). Following this step, the algorithm opts for the first option if \textit{headInitialFlag} is true, or the second option otherwise.

We illustrate \titleacronym{}$_3$ 
in Fig.~\ref{fig:manytomany-phrase}, where the target token \textit{\'ecrasement}, has two distinct source token options (\textit{october} (9) and \textit{crash} (11)). Both fall within the correct source phrase range (7-30). Since the source token \textit{october} already maps to \textit{octobre}, \textit{october} is removed from the source options for \textit{\'ecrasement}.

\vspace*{-.15in}
\subsubsection{$\titleacronym{}_3$: First-Come First-Assign (FCFA) SRL Projection.}

$\titleacronym{}_3$ is a new greedy FCFA SRL projection that transfers semantic roles using the remediated alignments (one-to-one mappings, \textit{remOneToOne}), as shown in Algorithm~\ref{alg:fcfaalgo_projection}. Alignments are provided as an input along with corresponding role labels transferred from English (\textit{srcSRLSet}).

\begin{wrapfigure}{r}{0.53\textwidth}
\vspace*{-.6in}
    \begin{minipage}{0.53\textwidth}
        \begin{algorithm}[H]
        \caption{First-Come First-Assign (FCFA) SRL Projection ($\titleacronym{}_3$)}
        \label{alg:fcfaalgo_projection}
        
        \scriptsize
        \vspace*{-.1in}
            \begin{flushleft}
            \textbf{Input} remOneToOne, srcSRLSet\\
            \textbf{Output} tgtSRLList\\
            \end{flushleft}
        \vspace*{-.15in}
        \begin{algorithmic}[1]
        \Function{FCFA}{remOneToOne, srcSRLSet}
            \State tgtSRLList $\gets[ ]$
            \For{$srcIdx, tgtIdx \in \textit{remOneToOne}$}
                \State \textit{srcSRL} $\gets$ \textit{srcIdx} th item of \textit{srcSRLSet}
                \If {$\textit{tgtIdx} \equiv \textit{eps}$}
                    \State \textit{tgtSRL} $\gets$ \textit{None}
                \Else
                    \State \textit{tgtSRL} $\gets$ \textit{srcSRL}
                    \State \textit{tgtSRLList}.insert((tgtIdx,tgtSRL))
                \EndIf
            \EndFor
            \State\Return tgtSRLList
        \EndFunction
        \end{algorithmic}
        \end{algorithm}
    \end{minipage}
 \vspace*{-.3in}
\end{wrapfigure}

Source side semantic roles are assigned to the remediated aligned target token (lines 3--9). Projection 
yields two outputs: a human interpretable alignment representation and a JSON formatted 
SRL representation. 
For example, in Fig.~\ref{fig:srl_projection} (a), token-level FCFA projects label (``O'') to \textit{octobre} and \textit{\'ecrasement} from \textit{october} and \textit{crash} (ordering). In addition, the source label from \textit{laptops} is projected to both \textit{ordinateurs} and \textit{portables}, leveraging the correct (one-to-many) alignment. Advancing beyond state-of-the-art (XSRL), many-to-one handling results in the retention of \textit{V} for \textit{effondrée} and the elimination of the hallucinated \textit{ARG4} for \textit{apart}.

\begin{wrapfigure}{r}{0.58\textwidth}
\scriptsize
\textbf{(a) Token-level FCFA SRL Projection} \\
\vspace*{-.2in}
\begin{flushleft}
\textbf{One-to-many}\\
\begin{tabular}{ l c l }
\textcolor{purple}{[O] laptop} & --- & \textcolor{blue}{[O] ordinateurs} \textcolor{black}{; 17-23} \\ 
\textcolor{purple}{[O] laptop} & --- & \textcolor{blue}{[O] portables} \textcolor{black}{; 17-24}\\ 
\end{tabular}
\end{flushleft}

\vspace*{-.25in}
\begin{flushleft}
\textbf{Many to one }\\
\begin{tabular}{ l c l }
\textcolor{purple}{[B-V] fall} & --- & \textcolor{blue}{[B-V] effondrée}\textcolor{black}{; 4-6}\\ 
\textcolor{purple}{\st{[B-ARG4] apart}} & --- & \textcolor{blue}{$\epsilon$}\\ 
\end{tabular}
\end{flushleft}
\vspace*{-.15in}
\begin{flushleft}

\vspace*{-.1in}
\begin{flushleft}
\textbf{Ordering }\\
\begin{tabular}{ l c l }
\textcolor{purple}{\st{[O] october}} & --- & \textcolor{blue}{$\epsilon$} \textcolor{black}{; 9-7}\\ 
\textcolor{purple}{[O] october} & --- & \textcolor{blue}{[O] octobre} 
\textcolor{black}{; 9-9}\\
\textcolor{purple}{[O] 1987} & --- & \textcolor{blue}{[O] 1987} 
\textcolor{black}{; 10-10}\\
\textcolor{purple}{[O] crash} & --- & \textcolor{blue}{[O] \'ecrasement} 
\textcolor{black}{; 11-7} 
\end{tabular}
\end{flushleft}

\textbf{(b) Phrase-level FCFA SRL Projection}
\\
\begin{tabular}{ l c l }
\textcolor{purple}{[I-ARG1] circuit} & --- & \textcolor{blue}{$\epsilon$} \\ 
\textcolor{purple}{[I-ARG1] breakers} & --- & \textcolor{blue}{[I-ARG1] disjoncteurs}\\
\textcolor{purple}{[B-V] installed} & --- & \textcolor{blue}{[B-V] install\'es}\\
\textcolor{purple}{[B-ARGM-TMP] after} & --- & \textcolor{blue}{[B-ARGM-TMP] apr\`es}\\
\textcolor{purple}{[I-ARGM-TMP] the} & --- & \textcolor{blue}{[I-ARGM-TMP] l'} \\ 
\textcolor{purple}{[I-ARGM-TMP] october} & --- & \textcolor{blue}{[I-ARGM-TMP] octobre} \\ 
\textcolor{purple}{[I-ARGM-TMP] 1987} & --- & \textcolor{blue}{[I-ARGM-TMP] 1987} \\ 
\textcolor{purple}{[I-ARGM-TMP] crash} & --- & \textcolor{blue}{[I-ARGM-TMP] écrasement} \\ 
\end{tabular}
\end{flushleft}

\vspace*{-.25in}
\caption{Token/phrase-level FCFA SRL projections}
\label{fig:srl_projection}
\vspace*{-.3in}
\end{wrapfigure}

Phrase-level projection operates similarly. In Fig.~\ref{fig:srl_projection} (b), \textit{october} aligns with \textit{octobre} and \textit{``ARGM-TMP''} transfers to \textit{octobre} (one-to-many). 
Due to the correct alignment of \textit{crash} with \textit{\'ecrasement}, \textit{``ARGM-TMP''} is transferred to \textit{\'ecrasement} (many-to-one). Furthermore, phrase-level projection considers whether the source language is head-initial or head-final. For example, \textcolor{purple}{\textit{[B-V-closed],  [B-ARGM-MNR-down]}} --- \textcolor{blue}{\textit{B-V-ferm\'e}}, \titleacronym{} projects \textit{``V''} from \textit{closed}, rather than \textit{``ARGM-MNR''} from \textit{down}.

\vspace*{-.2in}
\subsection{Explainability and Visualization}
\label{subsec:explain}
\vspace*{-.1in}
In contrast to blackbox LLMs, which 
do not elucidate the decisions behind language alignment and SRL projections, \titleacronym{} builds readily visualized representations that explain how it arrives at its output. 
Whereas prior work \cite{Hoffman2018MetricsProspects} has proposed metrics such as `goodness', `user satisfaction', and `understandability' as proxies for explainability, \titleacronym{} integrates human-interpretable representations directly into alignment and projection.

Two visualized products of our implementation (with French, Spanish as our test case) are: (a) a set of linguistically annotated alignment representations (one for each predicate indicated as ``V'') 
that provides a window into why/how the system produces its output while elucidating errors that can be readily remedied, as depicted in Fig.~\ref{fig:manytomany-phrase}; (b) a JSON formatted representation that specifies all semantic role-labeled tokens for each sentence, as depicted in Fig.~\ref{fig:flowchart} (\textit{French semantic role-labeled sentence}). 
These examples showcase our handling of hallucination remediation in the face of divergences and highlight the assignment of predicates and corresponding semantic roles on the target side.

\vspace*{-.2in}
\subsection{Model as a Diagnostic Tool}
\label{sec:diagnostic_tool}
\vspace*{-.07in}

\titleacronym{} employs a direct alignment-based source-to-target transfer mechanism, without requiring a filter or BERT Score (as implemented in XSRL). Moreover, the model based on this algorithm is an effective tool for assessing the accuracy of predicate and semantic role projection in longstanding community standard datasets. 
To illustrate this point, we explore a human-tagged English evaluation dataset from CoNLL-2009 \cite{ENSRC}, which has also been translated to French and Spanish data as part of XSRL's research \cite{Daza2020X-SRL:Dataset}. 

Preliminary tests using these datasets for SRL projection yield a much lower precision for \titleacronym{} than that of XSRL: \textit{\titleacronym{}: 65.9 (FR), 66.3 (ES), XSRL: 80.7 (FR), 85.4 (ES)}. Further investigation reveals that these data sets include a very large number of spurious V tags for non-predicates: 8341 (\titleacronym{}) vs. 3777 (XSRL), 8401 (\titleacronym{}) vs. 3870 (XSRL) for FR, ES, respectively. This is corroborated through analysis of part-of-speech (POS) attributes, which reveals that many verbs mislabeled as predicates do not have POS tag \textit{V} or \textit{VB(D)}.

This overabundance of incorrectly labeled non-predicates in the pre-existing English CoNLL-09 dataset (where spurious V-tagged tokens/phrases would more appropriately be labeled ARG0, ARG1, ARG2) leads to significantly corrupted projections. We thus leverage \titleacronym{} as a diagnostic tool, paving the way for refinements of the CoNLL-2009 gold dataset. We automatically remove the Y ($=$ Yes) flag for predicates that do not have part-of-speech \textit{V} or \textit{VB(D)}.\footnote{We have simplified the notion of \textit{predicate} considerably in this discussion, focusing on verbs; however, other parts of speech may serve as predicates. For example, \textit{destruction of the city} is a nominal phrase conveying a \textit{destroy} event with a single argument: \textit{the city}. Future work aims to explore other parts of speech as predicates.}  Correspondingly, incorrect transferal of falsely labeled predicates from the source is drastically reduced.

With this annotation refinement, we provide the updated new CoNLL-2009 dataset to the community. Correction of spurious predicate labels significantly improves the transferal of predicates and semantic roles during the application of \titleacronym{}. In  Section~\ref{sec:experiments}, all experiments use this newly updated dataset.

\vspace*{-.13in}
\section{Data and Experimental Setup}
\label{sec:experiments}
\vspace*{-.14in}



We use 
our updated English CoNLL-2009 data for projecting semantic roles to French and Spanish datasets. Human-validated FR/ES datasets, parallel to the EN-CoNLL, are provided by XSRL. The original CoNLL-2009 data incorporates semantic roles for headwords only. In our headword-level experiment, semantic roles from English headwords are projected to the headwords of the FR/ES datasets. Since phrase-level test datasets are unavailable, we employ AllenNLP's SRL-BERT 
to assign phrase-level semantic roles to the English corpora, which are then projected onto FR/ES corpora. 

Phrasal-level semantic role assignment further enhances the accuracy of SRL, ensuring phrasal coverage—a significant advance over the head-word labeling in the original resource. For instance, without our phrase-level enrichment, the word \textit{The} is considered a headword during SRL assignment in \textit{The Dow Jones industrials closed at 2569.26}. The result is a single, inappropriate semantic role assignment of ARG1 to the word \textit{The}. However, with our enrichment, an appropriate phrasal-level semantic role assignment is made possible: \textit{[ARG1- The Dow Jones industrials] [V-closed] [ARGM-EXT at 2569.26]}. This corrected output yields a more thorough, accurate representation, which is crucial for downstream tools such as those enumerated in Section~\ref{sec:intro}. 

French and Spanish corpora, including their semantic roles, are projected from 2046 English sentences using XSRL (see details in section~\ref{sec:relatedwork}) and \titleacronym{}. Subsequently, we evaluate these against both the community standard ground truth CoNLL-2009 (headword) from XSRL and the human judgment (phrasal). 
Our experiments run on 3 cores of AMD EPYC 75F3 32-Core Processor and using a NVIDIA A100 GPU.

\section{Results and Analyses}
\label{sec:results}
\vspace*{-.14in}

\begin{wraptable}{r}{.5\linewidth}
\vspace*{-.3in}
\fontsize{9}{10}
 \selectfont\caption{Word/Phrase-level projection evaluation for French and Spanish: \titleacronym{} vs. baseline (XSRL)}
\label{tab:GT_PH evaluation}
\vspace*{-.1in}
\begin{tabular}{|c|c|c|c|c|c|} \hline
\textbf{Model}& \textbf{Language}&\textbf{Level}  & \textbf{P} & \textbf{R} & \textbf{F1} \\ 
\hline
    XSRL&French & word  & 80.7 & 74.2 & 77.3\\ 
    \titleacronym{}&French & word  & 86.8 & 88.3 & \textbf{87.6} \\
    XSRL&Spanish & word  &  85.4 & 80.3 & 82.7 \\
    \titleacronym{}&Spanish & word  &  88.1 & 89.9 & \textbf{89.0} \\
    \hline
    XSRL&French & phrase & 91.9 & 74.4 & 82.2\\ 
    \titleacronym{}&French & phrase  & 98.9 & 81.1 & \textbf{89.1}  \\
    XSRL&Spanish & phrase  & 99.4 & 78.3  & 87.6 \\
    \titleacronym{}&Spanish & phrase  & 99.6  & 83.8 & \textbf{91.0}\\
\hline
    \end{tabular}
    \vspace*{-.24in}
\end{wraptable}

We explore the performance of two projection-based models: \titleacronym{} and XSRL. \titleacronym{} achieves higher F1 scores in comparison to XSRL on our test data in both word-level and phrasal-level (see Table~\ref{tab:GT_PH evaluation}). Performance improvements are obtained as well as explanability.

To evaluate the correctness of the French and Spanish projection outputs, we employ the ground truth data from CoNLL-2009 for the headword dataset. Linguistically trained human taggers proficient in French and Spanish evaluate the phrasal output. Both evaluations use precision (P), recall (R), and F1 scores. Thus, we have achieved explainable transferability of semantic roles more efficiently and with more accurate outputs (P, R, F1).

We evaluate \titleacronym{} against a human-validated CoNLL-2009 
that assigns semantic roles only to headwords, per the original XSRL algorithm. We compare XSRL and a variant of \titleacronym{} that produces only headword assignments against this same ground truth. In Table~\ref{tab:GT_PH evaluation}, \titleacronym{} projection to headwords outperforms XSRL, with an F1 of 77.3 vs. 87.6 (FR), 82.7 vs. 89.0 (ES). 

Furthermore, we conduct a post-analysis and evaluation of our phrasal-rich output against human judgment by French and Spanish proficient evaluators with linguistic training who evaluated 549 total labels (FR), and 582 total labels (ES). This analysis yields a F1 score (FR-89.1\%, ES-91.0\%, see Table~\ref{tab:GT_PH evaluation}). To our knowledge, this is the highest score achieved for this task, surpassing performance (accuracy) of single headword assignments without the overhead of human-labeled source data for French and Spanish. 

\vspace*{-.2in}
\section{Discussion: Beyond EN-FR / FR-ES}
\label{sec:discussion}
\vspace*{-.14in}
We explore hallucinations associated with linguistic divergences by considering language pairs beyond EN-FR / EN-ES.  
We consider Tagalog, a low-resource language notably influenced by Spanish at the word level~\cite{Baklanova2023SpanishNouns}, yet divergent from Spanish (and English) in that its subject follows the verb (VSO). Although our current study focuses on English as the source language, our future research focuses on Tagalog with both Spanish and English as source languages, further enriching our divergence exploration. This investigation aims to verify whether the pairs exhibit the divergent properties assumed by \titleacronym{} and to provide a framework for testing longstanding hypotheses about cross-language divergences in the context of alignment.

We introduce divergence metrics that 
count the number of misalignments on both the source and target sides. When the target language demonstrates a higher number of misalignments, this typically indicates a one-to-many divergence case. Conversely, when the source side yields more misalignment, this typically corresponds to a many-to-one case. \titleacronym{} effectively transfers semantic roles to the target language in both divergent cases, revealing the potential for generalizability to new language pairs.

As an early test case, we assess the applicability of our approach to English-Tagalog (EN-TL) or Spanish-Tagalog (ES-TL),\footnote{We use EN-ES-TL parallel data from LORELEI \cite{LORELEI}.} measuring misalignments on the source and target sides in both language pairs. Although mBERT alignment supports Tagalog, we are motivated to verify its effectiveness through this analysis, given that Tagalog is a low-resource language. We investigate alignment accuracy for EN-TL/ES-TL with the aid of a proficient human evaluator with ChatGPT \cite{chatgpt} support. On average, 3.27 words (21.69\%) and 4.04 words (26.77\%) per sentence are corrected in EN-TL and ES-TL alignment, respectively.

After applying this alignment correction, we measure the misalignment in the source and target sides, revealing a decrease in misalignment of the EN-TL (3.94 (22.77\%) to 3.0 (14.94\%) words on the source side, 4.54 (34.82\%) to 3.65 (26.05\%) words on the target side). Notably, the findings demonstrate comparable misalignment in both language pairs (EN-TL and ES-TL).

Alignment regeneration and correction are prerequisites
for employing \titleacronym{} for low-resource languages like Tagalog. Base alignment (mBERT) is insufficient for aligning Tagalog with other languages such as English or Spanish necessitating meticulous customization of the alignment for such low-resource cases. 

\vspace*{-.2in}
\section{Conclusions and Future Work}
\label{sec:conclusions}
\vspace*{-.14in}

We present a model for cross-language semantic role projection. Our work enhances semantically informed language processing with minimal overhead 
via a two-step process that rapidly identifies divergence cases and 
produces explainable, visualizable SRL output. We demonstrate performance improvements in accuracy without requiring a human-labeled French/Spanish corpus. Our evaluation relies on a community standard ground truth with SRL-tagged headwords (CoNNL-2009). Notable improvements are demonstrated when considering entire phrases, as evidenced by human judgments.

Future work will focus on expanding to other languages (Tagalog is underway) where hand-annotated labels are 
scarce. Although French and Spanish are investigated above, divergence-causing hallucinations, remediated by acknowledging the syntactic property of languages during \titleacronym{} have been noted across many other languages, e.g., Spanish (categorial; \cite{Dorr1994MachineSolution}), Korean (structural; \cite{Maniyar2021LinguisticPerceptive}), or German (light verb; \cite{Marzouk2021ChapterTranslation}).  
As such, it is expected that \titleacronym{} applies multilingually, both for mid-resource language pairs (e.g., English-Spanish/French) and for those that are low-resource language pairs (e.g., English-Tagalog).

Finally, our experiments reveal that a new model, \titleacronym{}, improves the multilingual SRL projection task. We provide French and Spanish corpora, including SRL information per predicates. 
Additionally, we utilize \titleacronym{} as a diagnostic tool to verify the accuracy of ground truth. Through this diagnostic tool, we identify errors in the data, enabling us to update and reproduce data for the language community. These data resources are not only beneficial for the SRL task but also may be leveraged for other tasks. 

\section*{Acknowledgements}
\vspace*{-.14in}
This research is based upon work supported by Defense Advanced Research Projects Agency (DARPA) under Contract No. HR001121C0186. Any opinions, findings and conclusions or recommendations expressed in this research are those of the authors and do not necessarily reflect the views of the US Government.
\vspace*{-.1in}



\bibliographystyle{splncs04}
\bibliography{references,custom.bib}


\end{document}